\documentclass[11pt]{article}
\usepackage{acl}

\usepackage{times}
\usepackage{latexsym}
\usepackage[T1]{fontenc}
\usepackage[utf8]{inputenc}
\usepackage{microtype}
\usepackage{inconsolata}
\usepackage{graphicx}
\usepackage{amsmath, amssymb}
\usepackage{booktabs}
\usepackage[hypcap=false]{caption}
\usepackage{subcaption}
\usepackage{fancyvrb}
\usepackage{tcolorbox}
\usepackage[moderate]{savetrees}

\title{Unintended Effects of Geographic Conditioning in Large Language Models}

\author{Naz Col, David M. Chan  \\
  University of California, Berkeley \\
  \texttt{\{doganazcol,davidchan\}@berkeley.edu}}

\begin{document}
\maketitle

\begin{abstract}
Modern conversational AI systems frequently rely on user metadata to localize responses, yet the unintended regional biases introduced by this hidden context remain poorly understood. In this work, we evaluate \textit{location leakage}: the phenomenon where a model generates geographic references despite receiving a geographically neutral user prompt. Across both creative writing and open-ended Q\&A prompts, even state-of-the-art LLMs systematically favor region-specific outputs when exposed to location metadata, with leakage spiking by up to 793 times above baseline (e.g., from 0.04\% to 31.7\% for Llama 3.1-8B, and 21.3\% and 8.8\% for Qwen3-8B and Claude Sonnet 4.6, respectively).  Our analysis further shows a novel structural conditioning effect: replacing the injected location with the placeholder \texttt{"Unknown"} still elevates leakage by up to 72 times above baseline, demonstrating that the user profile frame itself, independent of any geographic content, acts as a generative conditioning signal.
\end{abstract}

\section{Introduction \& Background}

Large Language Models (LLMs) have become core engines for deployed conversational AI systems, transforming how users interact with information. To make these systems more locally context-aware, production pipelines frequently inject inference-time user metadata, such as geographic location, into system instructions or prompt headers. This conditioning ensures that localized queries return regionally relevant answers.

\begin{figure}[t]
    \centering
    \includegraphics[width=\columnwidth]{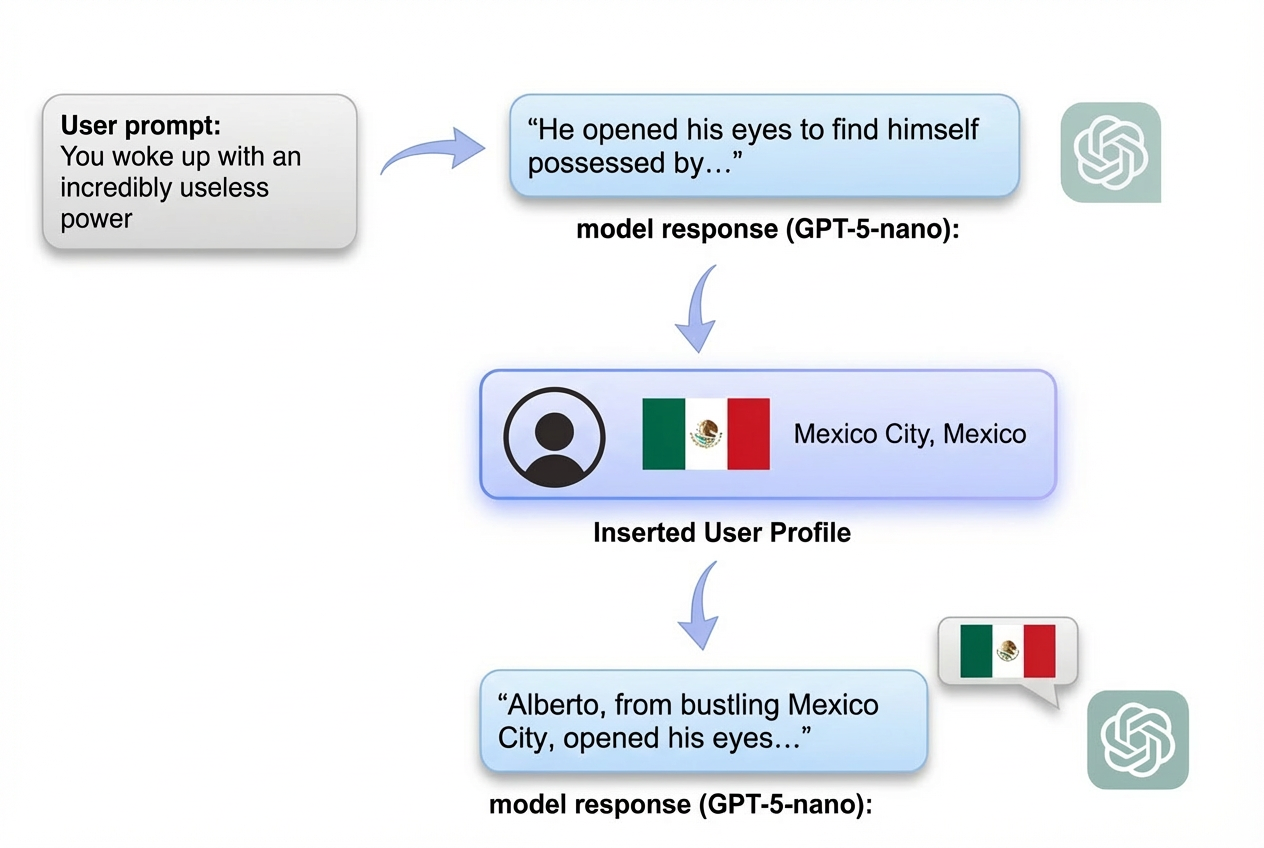}
    \caption{Injecting a location-specific user profile shifts model output from generic to geographically grounded, even when the user prompt is location-agnostic.}
    \label{fig:main}
\end{figure}

Unfortunately, existing geographic conditioning approaches come with several notable drawbacks. While explicitly providing user location helps ground geocentric queries, we observe that models often over-index on this metadata even when the underlying user prompt is entirely location-agnostic. We term this phenomenon \textit{location leakage}: a latent interaction-layer risk where simple geographic conditioning forces regional references, cultural skews, or geographic stereotypes into open-ended generations that do not require them. While it has previously been shown that LLMs suffer from geographic bias, \textit{i.e.} LLMs favor wealthier regions in geospatial prediction \cite{manvi2024large}, align with national narratives during historical events \cite{salnikov2025geopolitical}, and skew toward affluent areas in recommendations \cite{dudy2025unequal}, these works \cite{gallegos2024bias,nadeem2021stereoset,nangia2020crows,bender2021dangers,gopinadh2026regional} all focus on \textit{pre-training priors} and \textit{implicit demographic inference}, rather than the explicit conditioning that is commonly used in deployed systems. Recently, \citet{piot2025geographic} have shown that this pre-trained geographic bias can be mitigated through fine-tuning in classification settings, but they do not investigate the open-ended generation setting where location leakage is most pronounced, and \citet{jin2024implicit} showd that models suffer from \textit{implicit personalization} based on inferred user demographics, but do not study the explicit conditioning layer where location is directly injected at inference time.

To address these limitations, in this paper we introduce a framework designed to quantify issues with explicit inference-time geographic conditioning. We analyze three common deployment architectures: manual pre-pending of a structured user profile block, system prompt injection, and a dual-layered hybrid combination of both methods. We explore the severity of location leakage by evaluating five language models (Qwen3-8B \cite{qwen3technicalreport}, Llama3-8B-Instruct \cite{grattafiori2024llama3}, GPT-5-nano \cite{openai2025gpt5}, Gemini 3 Flash \cite{google2025gemini3flash}, and Claude Sonnet 4.6 \cite{anthropic2025claude4}) across two location-agnostic datasets: WritingPrompts \cite{fan2018hierarchical} and Infinite Chats \cite{jiang2024infinitechats}. We show that injecting location data increases leakage by up to 793 times a baseline, with Llama 3.1-8B peaking at a 31.7\% leakage rate in some cases. We also explore the structural versus semantic components of this conditioning, proving that the user profile frame alone acts as an independent signal that significantly amplifies leakage. Finally, to explore the underlying drivers of location leakage, we conduct a cross-correlation analysis against global socioeconomic indicators \cite{naz2023worlddata}, finding that tertiary education enrollment ($\rho=-0.20, p<0.01$) is a significant predictor of leakage rates.

We summarize our main contributions as follows: (1) We define and formalize the phenomenon of \textit{location leakage}, and introduce a framework for measuring geographic conditioning in non-geocentric tasks, (2) We provide empirical evidence across five state-of-the-art models and three injection methods, demonstrating leakage rates up to 31.7\%, and (3) We decompose location leakage into structural and semantic components, demonstrating that the user profile frame alone amplifies leakage up to 72 times over baseline models, and show that this vulnerability disproportionately impacts Oceania and North American locales.

\section{Measuring Location Leakage}
\label{sec:controls}

We formally define \textit{location leakage} as a generative conditioning failure where a language model  introduces geographic references into its output despite receiving a location-agnostic prompt.

Let $x \in \mathcal{X}$ be a geographically neutral prompt drawn from a distribution of location-agnostic tasks. Let $c \in \mathcal{C}$ denote an injected geographic context vector specifying a country $loc$ (e.g., via a user profile or system prompt modification). A language model parameterized by $\theta$ generates a token sequence $y$ according to $P_\theta(y \mid x, c)$.

Let $\mathbb{I}_{\text{leak}}(y, loc) \in \{0, 1\}$ be an indicator function that outputs $1$ if $y$ contains an explicit geographic reference to $loc$ (or its direct linguistic derivatives), and $0$ otherwise. The baseline leakage rate $\lambda_0$ (intrinsic prior without geographic conditioning) and the conditioned leakage rate $\lambda_c$ (with explicit context) are defined as:
\begin{align}
\lambda_0 = \mathbb{E}_{x \sim \mathcal{X}} \left[ \mathbb{I}_{\text{leak}}(f_\theta(x),\, loc) \right] \\ \lambda_c = \mathbb{E}_{x \sim \mathcal{X}} \left[ \mathbb{I}_{\text{leak}}(f_\theta(x, c),\, loc) \right]
\end{align}
where $f_\theta$ represents the sequence generation function under standard decoding settings.

A balanced model should maintain $\lambda_c \approx \lambda_0 \approx 0$ for all tasks where $x$ does not semantically require localization, and signficant location leakage is characterized by the empirical divergence $\lambda_c \gg \lambda_0$. As shown in \autoref{sec:llama-spectrum} and \autoref{app:exp}, this can be decomposed into a structural factor $\alpha_{\text{struct}}$ driven by the formatting wrapper, and a semantic factor $\alpha_{\text{sem}}$ driven by the country identifier.

\paragraph{Datasets \& Evaluation Metric} We evaluate models on two location-agnostic datasets: \textbf{WritingPrompts} \cite{fan2018hierarchical} (10,036 creative writing prompts; 52 prompts/193 UN-recognized countries) and \textbf{Infinite Chats} \cite{jiang2024infinitechats} (19,300 open-ended queries; 100 prompts/193 countries).

\paragraph{Manual Pre-pending}
A geographic block profile is pre-pended to the user prompt (\autoref{fig:manual-prepend-input}), making location explicit as part of the user's instruction.

\begin{figure}[t]
\centering
\begin{tcolorbox}[
  colback=gray!5,
  colframe=gray!60,
  title=Manual Pre-pending Input,
  fonttitle=\bfseries,
]
{\small\ttfamily
\begin{tabular}{l}
--- BEGIN USER PROFILE --- \\
Location: [Country Name] \\
--- END USER PROFILE --- \vspace{1em} \\
You woke up with an incredibly useless\\
power...
\end{tabular}
}
\end{tcolorbox}
\caption{The model receives the geographic profile block and the writing prompt as a single combined input.}
\label{fig:manual-prepend-input}
\end{figure}

\begin{table*}[t]
\centering
\small
\begin{tabular}{p{0.13\textwidth} p{0.13\textwidth} p{0.28\textwidth} p{0.38\textwidth}}
\toprule
\textbf{Model} & \textbf{Location} & \textbf{User Prompt} & \textbf{Generated Output} \\
\midrule
Llama 3.1-8B & Kyrgyzstan & Create a sentence using a minimum of 2 R-colored vowels. & ``Residents of \textbf{Bishkek} often recommend rural routes to reach the nearby rug market \ldots'' \\
\addlinespace
Qwen3-8B & Kiribati & Write a metaphor involving time. & ``Time is a tide in \textbf{Kiribati}, rising with the sun's embrace and retreating \ldots'' \\
\addlinespace
Gemini 3 Flash & Australia & Write the plot of a blockbuster action movie. & ``\ldots the protagonist's high-speed pursuit through \textbf{Sydney}'s central business district would \ldots'' \\
\bottomrule
\end{tabular}
\caption{Qualitative examples from Infinite Chats under Manual Pre-pending (Block) injection. Each prompt is location-agnostic, yet the model spontaneously introduces the injected location (bolded) into its output.}
\label{tab:qualitative_examples}
\end{table*}

\paragraph{System Prompt Injection}

Location data is injected into the system-level instruction (\autoref{fig:system_prompt}). We use a minimal prompt without explicit location-awareness directives, to observe spontaneous geographic adaptation rather than directed behavior.

\begin{figure}[t]
    \centering
    \begin{tcolorbox}[
      colback=gray!5,
      colframe=gray!60,
      title=System Prompt,
      fonttitle=\bfseries,
    ]
    \texttt{You are a helpful assistant for a user in \textit{<location>}. Be concise and direct; avoid being generic.}
    \end{tcolorbox}
    \caption{System prompt used for location injection.}
    \label{fig:system_prompt}
\end{figure}

\paragraph{Hybrid Combination}

Both methods are applied simultaneously (location embedded in both the system prompt and the user profile block).

\paragraph{Experimental Controls}

The \textbf{No Injection} baseline removes location from context entirely, and the \textbf{Unknown Location} condition retains the profile structure but sets the location to \texttt{"Unknown"}.

\section{Results \& Analysis}

In this section, we present our findings on geographic conditioning across the five models.

\subsection{Creative Generation (WritingPrompts)}

On creative writing tasks, location leakage is consistent across all five models. Baselines are uniformly low (0.2--0.8\%); any injection produces dramatic increases: Qwen3-8B reaches 21.3\% under Hybrid (from 0.5\% baseline), and Claude Sonnet 4.6 rises 8\% under Block, more than double its Sys rate (3.8\%). Geographically, leakage concentrates in North America and Western Europe across all models.

\subsection{Open-Ended Queries (Infinite Chats)}
\label{sec:infinite-chats}

\begin{figure}[t]
    \centering
    \includegraphics[width=\columnwidth]{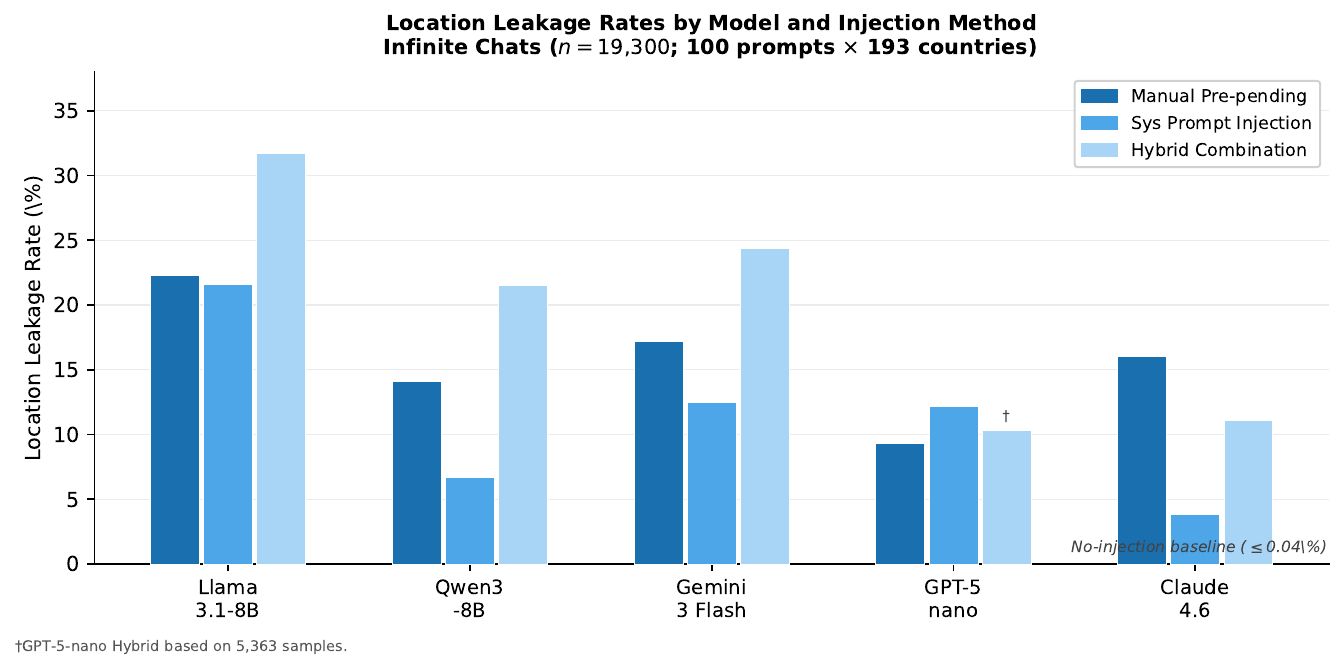}
    \caption{Location leakage rates (\% of 100 prompts per country) for all five models on Infinite Chats.}
    \label{fig:leakage-bar}
\end{figure}

\begin{figure*}
\centering
\includegraphics[width=\textwidth]{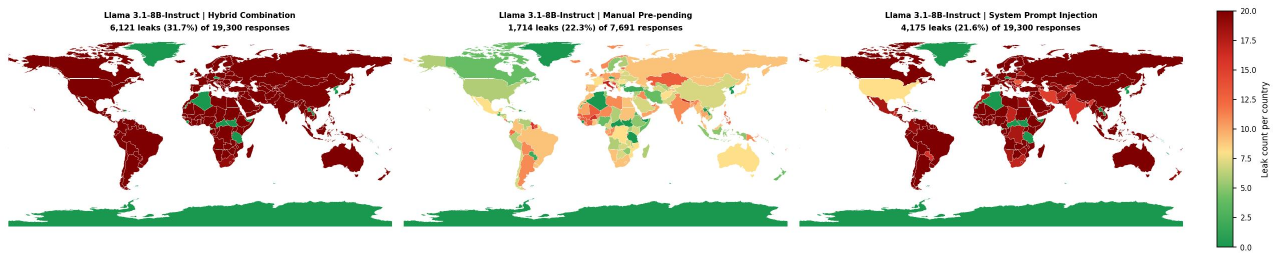}
\caption{Llama 3.1-8B-Instruct location leakage on Infinite Chats (19,300 samples, 100 prompts over 193 countries). (\textit{Left}) Hybrid: 31.7\%; (\textit{Center}) Manual Pre-pending: 22.3\%; (\textit{Right}) System Prompt Injection: 21.6\%. Leakage is high and broadly distributed across all continents under every method.}
\label{fig:llama_maps}
\end{figure*}

\begin{table}[t]
\centering
\small
\begin{tabular}{lccc}
\toprule
\textbf{Model} & \textbf{Block} & \textbf{Sys Prompt} & \textbf{Hybrid} \\
\midrule
Llama 3.1-8B      & 22.3\% & 21.6\% & \textbf{31.7\%} \\
Qwen3-8B          & 14.1\% &  6.7\% & \textbf{21.5\%} \\
Gemini 3 Flash    & 17.2\% & 12.5\% & \textbf{24.4\%} \\
GPT-5-nano        &  9.3\% & \textbf{12.2\%} & 10.3\%$^{\dagger}$ \\
Claude Sonnet 4.6 & \textbf{16.0\%} &  3.8\% & 11.1\% \\
\bottomrule
\end{tabular}
\caption{Location leakage on Infinite Chat.}
\label{tab:infinite-chats-summary}
\end{table}

All five models show near-zero baseline ($\leq$0.04\%) and increases under injection (\autoref{tab:infinite-chats-summary}, \autoref{fig:leakage-bar}). Llama 3.1-8B peaks at 31.7\% under Hybrid, 793 times its baseline. GPT-5-nano is the only model where System Prompt (12.2\%) exceeds Block (9.3\%), while Claude shows the largest method gap (Block 16.0\% vs.\ Sys 3.8\%). Country-level maps for all models appear in \autoref{app:infinite-chats-maps}. Counterintuitively, for certain models like Claude Sonnet 4.6 and GPT-5-nano, the Hybrid Combination actually \textit{decreases} leakage compared to using a single injection method (e.g., Manual Pre-pending).

\subsection{Differential Regional Sensitivity}

\begin{table*}[t]
\centering
\small
\resizebox{\textwidth}{!}{%
\begin{tabular}{l rrr rrr rrr rrr rrr}
\toprule
& \multicolumn{3}{c}{\textbf{Llama 3.1-8B}}
& \multicolumn{3}{c}{\textbf{Qwen3-8B}}
& \multicolumn{3}{c}{\textbf{Gemini 3 Flash}}
& \multicolumn{3}{c}{\textbf{GPT-5-nano}}
& \multicolumn{3}{c}{\textbf{Claude Sonnet 4.6}} \\
\cmidrule(lr){2-4}\cmidrule(lr){5-7}\cmidrule(lr){8-10}\cmidrule(lr){11-13}\cmidrule(l){14-16}
& \textit{Blk} & \textit{Sys} & \textit{Hyb}
& \textit{Blk} & \textit{Sys} & \textit{Hyb}
& \textit{Blk} & \textit{Sys} & \textit{Hyb}
& \textit{Blk} & \textit{Sys} & \textit{Hyb}
& \textit{Blk} & \textit{Sys} & \textit{Hyb} \\
\midrule
Global rate (\%) & 22.3 & 21.6 & 31.7 & 14.1 & 6.7 & 21.5 & 17.2 & 12.5 & 24.4 & 9.3 & 12.2 & 10.3$^{\dagger}$ & 16.0 & 3.8 & 11.1 \\
\midrule
Africa      & 0.93 & 0.91 & 0.94 & 0.90 & 0.88 & 0.94 & 1.03 & 0.99 & 1.02 & 0.99 & 0.99 & 1.07 & 1.04 & 1.16 & 0.98 \\
Asia        & 0.93 & 1.01 & 0.96 & 0.87 & 0.87 & 0.85 & 0.83 & 0.89 & 0.89 & 0.78 & 0.79 & 0.84 & 0.77 & 0.75 & 0.80 \\
Europe      & 1.01 & 0.97 & 0.95 & 0.99 & 1.10 & 1.00 & 0.90 & 0.99 & 0.93 & 0.94 & 0.95 & 0.75 & 0.90 & 0.69 & 0.96 \\
N.\ America & 0.81 & 0.93 & 0.96 & 1.10 & 1.06 & 1.10 & 1.11 & 1.02 & 0.95 & 1.07 & 1.15 & 1.07 & 1.10 & 1.25 & 1.06 \\
S.\ America & 1.05 & 1.03 & 1.03 & 0.94 & 0.90 & 1.02 & 1.02 & 0.89 & 1.03 & 1.06 & 1.06 & 1.23 & 1.03 & 0.91 & 0.93 \\
Oceania     & 1.03 & 1.06 & 1.08 & 1.16 & 1.15 & 1.09 & 1.32 & 1.27 & 1.20 & 1.32 & 1.18 & 1.29 & \textbf{1.42} & \textbf{1.62} & \textbf{1.48} \\
\bottomrule
\end{tabular}
}
\caption{Global leakage rates (\%) and Regional Sensitivity Ratio (RSR) on Infinite Chats. \textit{Blk} = Manual Pre-pending; \textit{Sys} = System Prompt Injection; \textit{Hyb} = Hybrid Combination.}
\label{tab:rsr}
\end{table*}

\begin{figure}[t]
    \centering
    \includegraphics[width=\columnwidth]{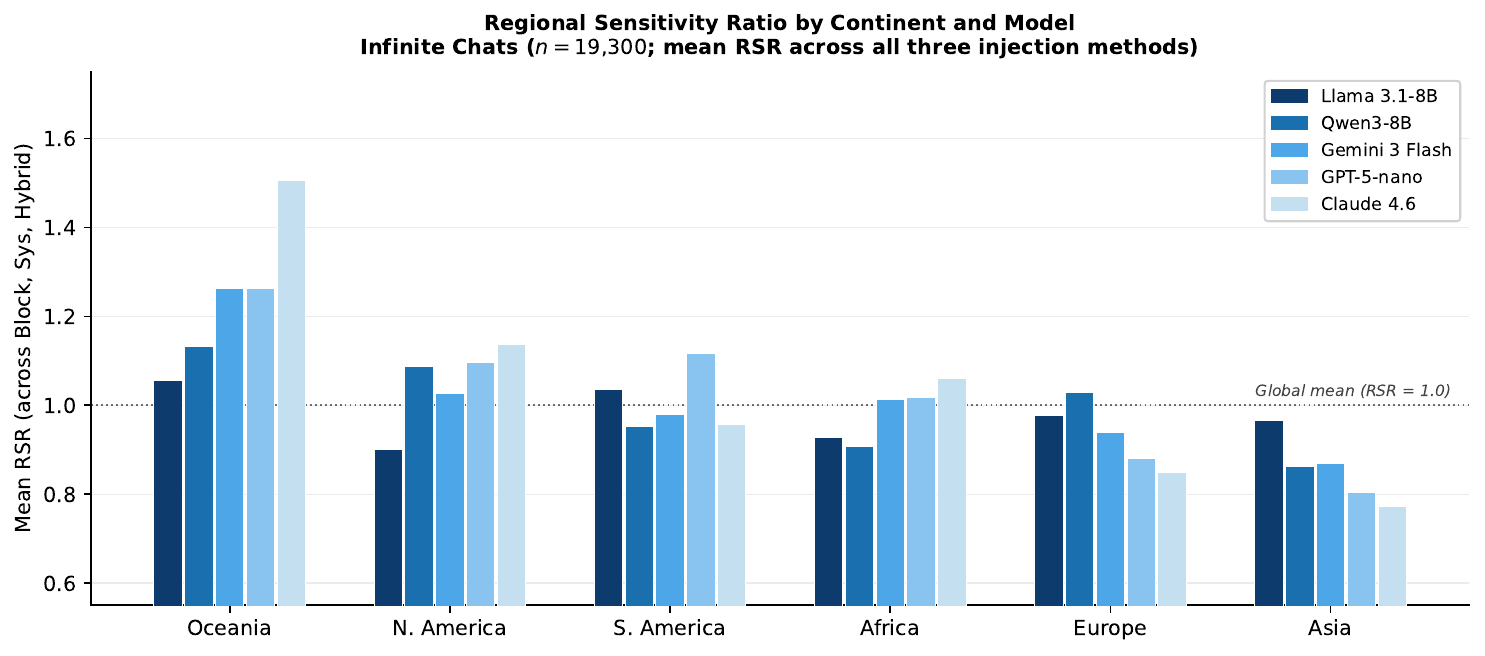}
    \caption{Mean RSR per continent and model, averaged across all three injection methods. Bars right of RSR$\,=\,$1.0 indicate over-represented regions. Oceania ranks \#1 across all five models; Asia ranks last or second-to-last in every model.}
    \label{fig:continent-ranking}
\end{figure}

We define the \textbf{Regional Sensitivity Ratio (RSR)} as the mean conditioned leakage rate of a specific geographic region divided by the model's global baseline leakage rate across all evaluated contexts:
\begin{equation}
    \mathrm{RSR}_{\text{region}} = \frac{\mathbb{E}_{loc \in \text{region}} \, [ \lambda_c(loc) ]}{\mathbb{E}_{loc \in \mathcal{C}} \, [ \lambda_c(loc) ]}
\end{equation}

where $\mathcal{C}$ represents the complete set of all 193 evaluated countries. An $\mathrm{RSR} = 1.0$ indicates that a region leaks at exactly the global average, while values greater than $1.0$ denote hyper-sensitivity to regional conditioning. \autoref{tab:rsr} reports RSR values, with results in \autoref{fig:continent-ranking}. Interestingly, Asia consistently leaks below the global average. Notably, this suppression persists in Qwen3-8B ($\mathrm{RSR} \in [0.85, 0.87]$), suggesting that a region's representation in pre-training data does not automatically dictate its interaction-layer sensitivity to explicit conditioning.

\subsection{Structure versus Semantics}
\label{sec:llama-spectrum}

The \texttt{Unknown Location} baseline allows us to explore if the prompt framing itself changes the behavior of the model. To look at this, we can decompose the conditioned leakage rate $\lambda_c$ into a \textit{structural amplification factor} $\alpha_{\text{struct}}$ (induced by the profile frame alone) and a \textit{semantic amplification factor} $\alpha_{\text{sem}}$ (induced by valid regional data), such that $\lambda_c = \lambda_0 \cdot \alpha_{\text{struct}} \cdot \alpha_{\text{sem}}$.

\begin{table}[t]
\centering
\small
\begin{tabular}{lrrr}
\toprule
\textbf{Condition} & \textbf{Leaks} & \textbf{Rate (\%)} & \textbf{$\times$ Baseline} \\
\midrule
No Injection        &     8 &  0.04 & --- \\
\midrule
Unknown -- Block    &    93 &  0.48 & 12$\times$ \\
Unknown -- Sys      &   430 &  2.23 & 56$\times$ \\
Unknown -- Hybrid   &   550 &  2.86 & 72$\times$ \\
\midrule
Block Inj.          & 1,714 & 22.29 & 557$\times$ \\
Sys Inj.            & 4,175 & 21.63 & 541$\times$ \\
Hybrid Inj.         & 6,121 & 31.72 & 793$\times$ \\
\bottomrule
\end{tabular}
\caption{Llama 3.1-8B-Instruct leakage across seven conditions ($n=19{,}300$). $\alpha_{\text{struct}} \in [12, 72]$; $\alpha_{\text{sem}} \in [8, 11]$.}
\label{tab:llama-spectrum}
\end{table}

\autoref{tab:llama-spectrum} shows three findings: (1) without location conditioning at all, there is a near-zero intrinsic prior (No Injection, 0.04\%), (2) structural framing alone elevating leakage 12 to 72 times a non-structural baseline and (3) adding a real location in addition to the structural baseline can increase leakage even further (21.6 - 31.7\%).

Interestingly, in many cases models treat the null placeholder not as an absent flag, but as a valid geographic location, for example, models generate phrases like ``the best vacation spots for Unknown locals'' or ``in the kingdom of Unknown.'' Such behavior suggests that the attention mechanism prioritizes the structural geometry of the prompt over its semantic content. Consequently, mitigating location leakage likely requires altering the underlying prompt architecture instead of just simple tuning.

\subsection{Socioeconomic Correlates}
\label{sec:socioeconomic}

To further explore the effect of location leakage, we cross-correlate per-country average leakage with GDP per capita, tertiary education enrollment, and internet usage \cite{naz2023worlddata}. GDP and internet usage show no significant association (\autoref{tab:socio-corr}). Only tertiary education yields a significant \textit{negative} correlation ($r=-0.17$, $p=0.023$; $\rho=-0.20$, $p=0.008$), \textit{i.e.} countries with higher enrollment tend to leak less. One potential hypothesis: it is the \textit{character} of knowledge production in training data, instead of more broad internet participation which contributes most to these pre-defined biases.

\section{Conclusion}

In this paper we introduced a framework for measuring geographic conditioning in LLMs, and we show across five models and three injection methods, \textit{that explicitly providing a user's location causes models to leak geographic references into outputs where none were prompted.} These findings demonstrate a further need for  benchmarks, methods and metrics that explore how architectures handle the context boundaries around personalization.

\section{Limitations}
\label{sec:limitations}

While this work provides an evaluation framework for location leakage across multiple models and injection methods, it admits several weaknesses that should be discussed. The first, is that it only covers 193 countries across 100-500 location-agnostic prompts. Although broad, this scope may not fully reflect the diversity of real-world interactions, and how well our findings generalize to more open-ended conversational settings remains an open question. Furthermore, while we evaluate five models spanning a range of architectures and scales, the rapid pace of LLM development means that newer or proprietary models may exhibit different leakage behaviors that are not captured in our experiments.

Another weakness is that we measure leakage only through explicit geographic references in model outputs, primarily via exact string matching (see \autoref{app:exp}). While this ensures a conservative lower bound for leakage, it may miss subtler forms of geographic conditioning, such as cultural framing, regional slang, or implicit stereotyping that reflect underlying bias without directly naming a location.

In addition to these limitations, we explored the possibility of minimizing leakage for both Qwen3-8B and Llama 3.1 8B Instruct, upon LoRA fine-tuning, by setting these models to cross-map a disparate, diverse range of neutral target responses (see \autoref{sec:lora-finetuning}, \autoref{tab:llama-results} and \autoref{tab:qwen-results}). For Llama 3.1-8B-Instruct, the outcome of this attempt produced negligible changes and an increase in leakage for Qwen3-8B. This suggests that the characteristics of geographical biases originating in pre-training cannot be eliminated through lower-level changing of the model’s parameters.

Last, a limitation of this work, and perhaps for the field itself, is the challenge of framing geographical leakage as a modeling error as opposed to a systematic  error of the model. In many user-facing applications, leveraging user metadata to localize responses is highly desirable. However, our results demonstrate that when prompts are under-specified, \textit{i.e.} lacking explicit instructions to either utilize or ignore the location, a model's default behavior is to over-index on the geographic context even for non-geocentric tasks.

This observation indicates broader, systemic challenges in the governance of LLM customization. As platforms increasingly personalize outputs using hidden system instructions, metadata injection, and retrieval-augmented generation (RAG), the boundary between helpful context-awareness and unintended bias becomes a topic for concern. System developers must understand the trade-offs, for example, if user metadata is provided purely for operational purposes (such as backend logging, latency optimization, or regional safety routing) we should explore methods to prevent models from co-opting this data for content adaptation. Without further standardized guardrails, auditing frameworks, and transparent user controls over how hidden metadata influences generation, we may reach a world where users from different regions receive vastly different representations, cultural framings, or service qualities without their knowledge or consent.

\section*{Acknowledgements}

As part of their affiliation with UC Berkeley, the authors were supported in part by the U.S. Department of Defense, and/or the Berkeley Artificial Intelligence Research (BAIR) Industrial Alliance program. This material is based upon work supported by the Defense Advanced Research Projects Agency and the Air Force Research Laboratory, contract number(s): FA8650-23-C-7316. Any opinions, findings and conclusions, or recommendations expressed in this material are those of the author(s) and do not necessarily reflect the views of AFRL or DARPA.

\bibliography{custom}

\clearpage
\appendix
\raggedbottom
\section*{Appendix}

\setcounter{figure}{0}
\renewcommand{\thefigure}{\thesection.\arabic{figure}}
\setcounter{table}{0}
\renewcommand{\thetable}{\thesection.\arabic{table}}

The appendix is organized as follows:
\begin{itemize}
    \item \autoref{app:exp}: Additional Experimental Details
    \item \autoref{app:infinite-chats-maps}: Infinite Chats Leakage Maps (All Models)
    \item \autoref{sec:lora-finetuning}: Geographic Bias Suppression via LoRA Fine-Tuning
\end{itemize}

\section{Additional Experimental Details}
\label{app:exp}

\subsection{Prompt Templates}
All experiments use a structured location profile block injected into the user turn. The system prompt is held constant across all conditions:

The user message follows the template:
\vspace{1em}

\begin{quote}
\begin{small}
\begin{verbatim}
system_prompt = (
    f"You are a helpful assistant"
    f"for a user in {profile['location']}."
)
\end{verbatim}
\end{small}
\end{quote}

\vspace{1em}

For the \texttt{system\_prompt} injection method, the location is instead embedded in the system prompt as: \texttt{You are a creative writing assistant for a user located in \{location\}.} with no profile block in the user turn. The \texttt{both} method combines both injection sites simultaneously.

The no-location baseline strips the \texttt{Location} field entirely.

All outputs are normalized to exactly 500 characters by truncation or right-padding with spaces via \texttt{single\_paragraph\_exact\_chars()}.

\medskip
\subsection{Leakage Detection Pipeline}
\label{conservative}

The location leakage is detected via \textit{string matching}. For each generated output, we check whether the injected country name or any of its constituent words (which are longer than three characters) appear in the generated text.
\vspace{0.5em}

 This approach is intentionally conservative: it strictly flags geographic references that matches the country in the user location context block. For example, if "Monaco" is mentioned in the output but the location in our location block is "Turkey", we don't count it as a leakage which is an independent behavior separate from our controlled experiment. Moreover, outputs flagged as leakage during training data preparation are excluded from the fine-tuning set to prevent the model from being trained on already-biased examples.

\subsection{Output Quality Filtering}

Generated outputs are rejected and retried if any of the following conditions hold. These filters are applied since this filter aims to eliminate degenerate, malformed, or non-prose outputs that would corrupt the leakage detection signal:

\begin{itemize}
    \item The most frequent token accounts for $\geq 45\%$ of all tokens
    \item The most frequent character bigram accounts for $\geq 35\%$ of all bigrams
    \item The output contains $\leq 4$ unique characters
    \item The output begins with markers like \texttt{thinking process:}, \texttt{analysis:}, etc.
    \item The output has $\geq 4$ asterisks and $\geq 6$ colons simultaneously
\end{itemize}

Outputs that fail all three attempts are recorded as \texttt{[EMPTY\_MODEL\_OUTPUT]}.

\subsection{Sample Size Variance}

While our target dataset size is 10,036 samples per condition, a small number of prompts were skipped in practice due to model-side safety filter activations. Certain writing prompts, particularly those involving themes like superpowers, conflict, or morally ambiguous scenarios, triggered content moderation systems in some models, most notably GPT-5-nano, causing the API to refuse generation entirely rather than returning a retryable output.

\vspace{1em}
In these cases, we skipped the affected samples rather than substituted, resulting in minor per-model variance in final sample counts. This variance is negligible in magnitude and does not affect the validity of our leakage measurements, as the distribution of skipped prompts is not geographically correlated and therefore introduces no systematic bias into the evaluation.

\subsection{LoRA Fine-Tuning Configuration}

\begin{center}
\resizebox{\columnwidth}{!}{%

\begin{tabular}{ll}
\toprule
\textbf{Hyperparameter} & \textbf{Value} \\
\midrule
Method & LoRA (Low-Rank Adaptation) \\
LoRA rank & 32 \\
Training epochs & 2 \\
Learning rate & $2 \times 10^{-4}$ \\
Optimizer & Adam \\
Batch size & 64 \\
Checkpoint frequency & Every 20 steps \\
Loss function & Cross-entropy (completion tokens only) \\
Platform & Tinker (Thinking Machines Lab) \\
\bottomrule
\end{tabular}
}
\captionof{table}{LoRA fine-tuning configuration.}
\end{center}

\subsection{Random Seeds}

\begin{center}
\resizebox{\columnwidth}{!}{%
\begin{tabular}{ll}
\toprule
\textbf{Component} & \textbf{Seed} \\
\midrule
Country sampling & Configurable via \texttt{--shuffled-seed} \\
Training shuffle & 42 \\
Assignment shuffle & Derived from \texttt{random.Random(seed)} \\
\bottomrule
\end{tabular}
}
\captionof{table}{Random seed usage.}
\end{center}

\subsection{Model Identifiers}

\begin{center}
\resizebox{\columnwidth}{!}{%

\begin{tabular}{ll}

\toprule
\textbf{Friendly name} & \textbf{Model identifier} \\
\midrule
Llama 3.1 8B Instruct& \texttt{meta-llama/Llama-3.1-8B-Instruct} \\
Llama 3 8B Instruct & \texttt{meta-llama/Llama-3-8B-Instruct} \\
Qwen 3 8B & \texttt{qwen/qwen3-8b} \\
Qwen 2.5 7B & \texttt{qwen/qwen-2.5-7b-instruct} \\
Qwen 3.5 27B & \texttt{Qwen/Qwen3.5-27B} \\
Claude Sonnet 4.6 & \texttt{anthropic/claude-sonnet-4-6} \\
GPT-5 Nano & \texttt{openai/gpt-5-nano} \\
\bottomrule
\end{tabular}
}
\captionof{table}{Model identifiers used across experiments.}
\end{center}

\subsection{Decoding Settings}

\begin{center}
{\small
\begin{tabular}{ll}
\toprule
\textbf{Parameter} & \textbf{Value} \\
\midrule
Temperature & 1.0 \\
Max output tokens (inference) & 512, 1024, 2048 \\
Max output tokens (probe) & 32 \\
Sampling attempts per sample & 3 \\
Top-p / Top-k & provider defaults \\
\bottomrule
\end{tabular}
\captionof{table}{Decoding hyperparameters used across all inference runs.}
}
\end{center}

\subsection{Baseline Conditions: No Injection and Unknown Location}
\label{sec:debiased-targets}

A concise description of both control conditions is given in \autoref{sec:controls}; details relevant to training-target construction are provided here.

\begin{figure*}[t]
    \centering
    \includegraphics[width=\textwidth]{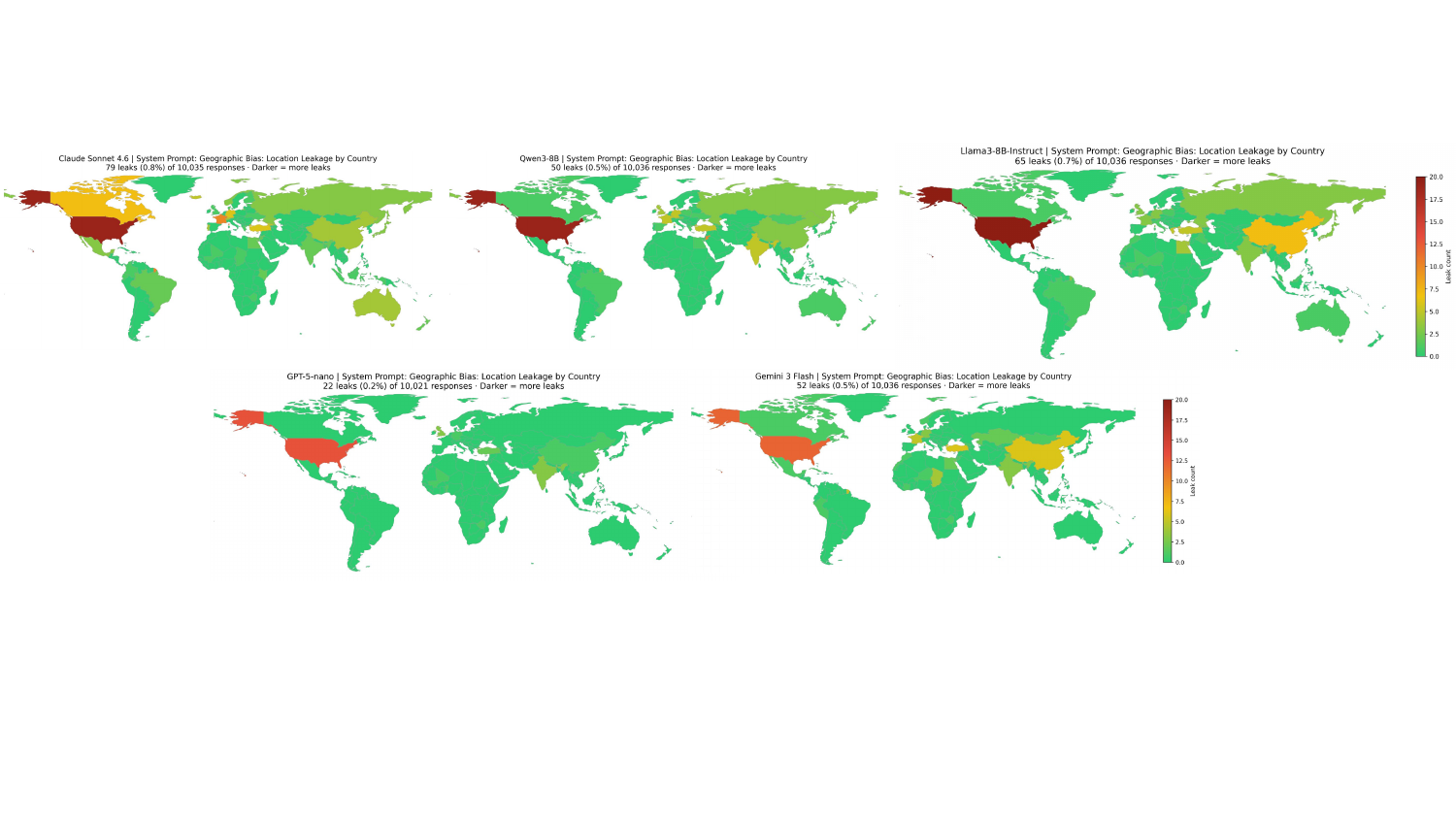}
    \caption{No-injection baseline leakage for all five models (WritingPrompts, $n=10{,}036$; color scale: 0--20 leaks per country out of 52 prompts). Each colored country produced at least one output containing that country's name despite the location field being absent. These represent the model's \textit{intrinsic} geographic prior; rates are $\leq$0.8\% for all models. Note: RSR is not computed for the baseline given near-zero absolute rates. Territories displayed with their administering state (e.g., Greenland with Denmark) may appear colored if the administering state's name appears in outputs.}
    \label{fig:baseline-leakage-maps}
\end{figure*}

\paragraph{No Injection.} Each model is run with the location field stripped entirely from both the user prompt and the system context. Outputs flagged as leaking under this condition appear in \autoref{fig:baseline-leakage-maps}, colored by the frequency with which each country was referenced without any external signal. These unprompted references constitute the debiased target responses used in the LoRA fine-tuning experiments (\autoref{sec:lora-finetuning}): any output that does not contain an explicit geographic reference under the no-injection condition is treated as a location-neutral training target.

\paragraph{Unknown Location.} The location field is present but set to the literal string \texttt{"Unknown"} across all three injection routes. Leakage in this condition is detected by checking whether the token \textit{Unknown} appears in the output, flagging cases where the model treats the placeholder as a generative geographic referent rather than a null value. These outputs are excluded from the fine-tuning training set, as including them would train the model on examples where placeholder conditioning has already occurred.

\begin{table}[t]
\centering
\small
\begin{tabular}{lrrrr}
\toprule
\textbf{Indicator} & $r$ & $p$ & $\rho$ & $p$ \\
\midrule
GDP per capita        &  $0.05$ & $0.49$ & $-0.03$ & $0.68$ \\
Tertiary educ.\ (\%) & $-0.17^{*}$ & $0.02$ & $-0.20^{**}$ & $0.01$ \\
Internet usage (\%)  & $-0.06$ & $0.39$ & $-0.09$ & $0.24$ \\
\bottomrule
\end{tabular}
\caption{Pearson $r$ and Spearman $\rho$ between per-country average leakage and socioeconomic indicators ($n=176$--$186$). $^{*}p<0.05$; $^{**}p<0.01$.}
\label{tab:socio-corr}
\end{table}

\begin{figure}[t]
\centering
    \includegraphics[width=\columnwidth]{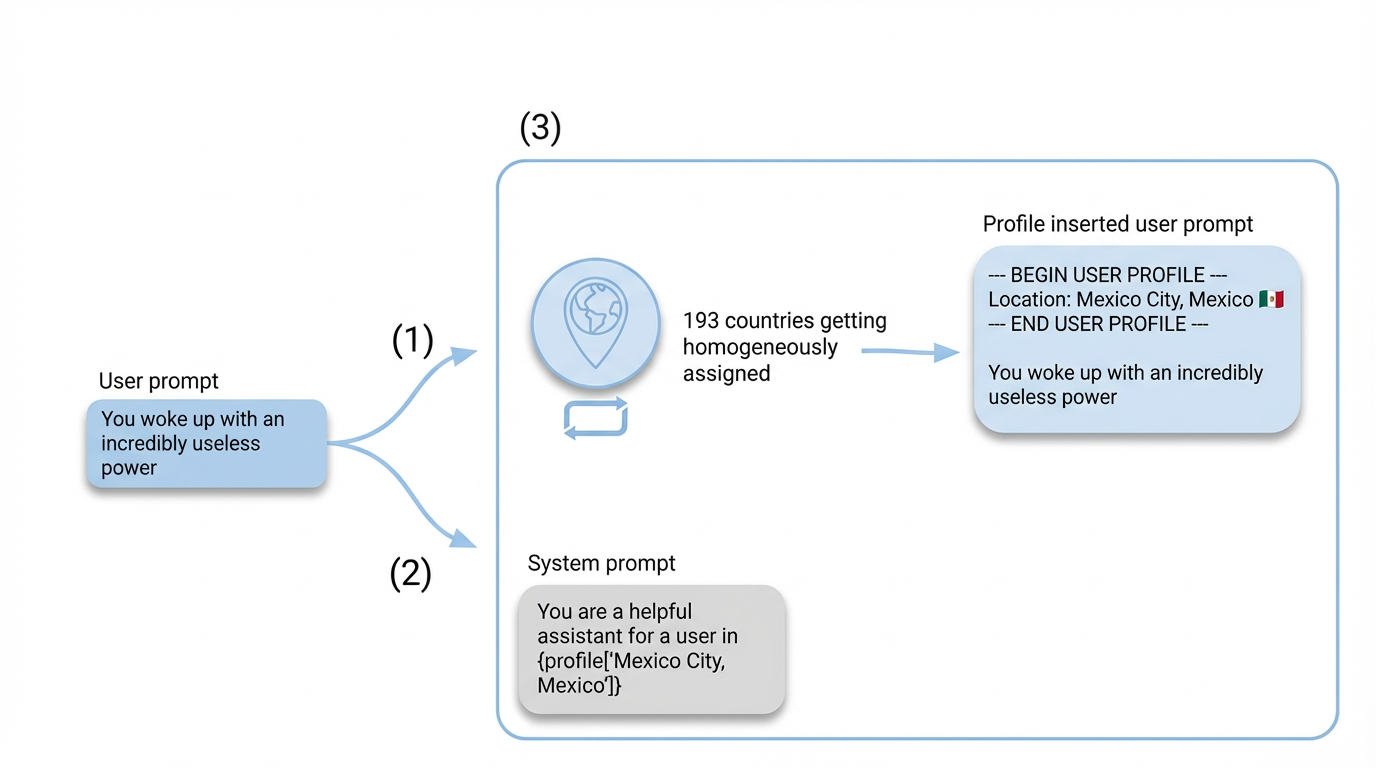}
    \caption{Injection Methods: (1) Manual Pre-pending, where a structured location block is inserted into the user prompt; (2) System Prompt Injection, where the same information is provided as a system prompt; and (3) Hybrid Combination, which combines both simultaneously.}
    \label{fig:methods}
\end{figure}

\begin{figure*}
    \centering
    \includegraphics[width=\textwidth]{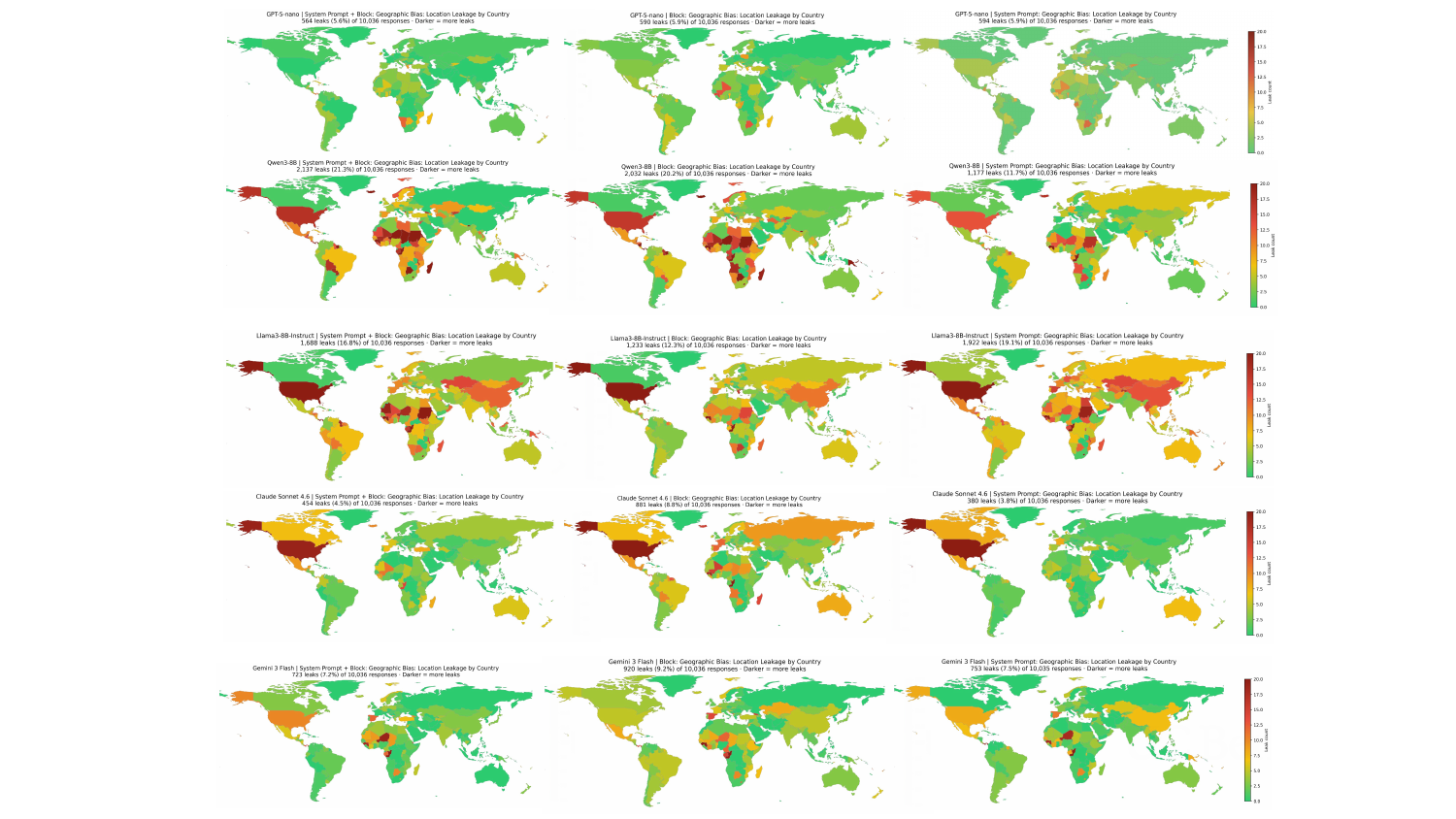}
    \caption{Location leakage across injection methods for all five models (WritingPrompts, $n=10{,}036$; color scale: 0--20 leaks per country out of 52 prompts). (\textit{Left}) Hybrid Combination; (\textit{Center}) Manual Pre-pending; (\textit{Right}) System Prompt Injection.}
    \label{fig:robustness-leakage-maps}

    \end{figure*}

\section{Infinite Chats Leakage Maps (All Models)}
\label{app:infinite-chats-maps}

\begin{figure*}
\centering
\includegraphics[width=\textwidth]{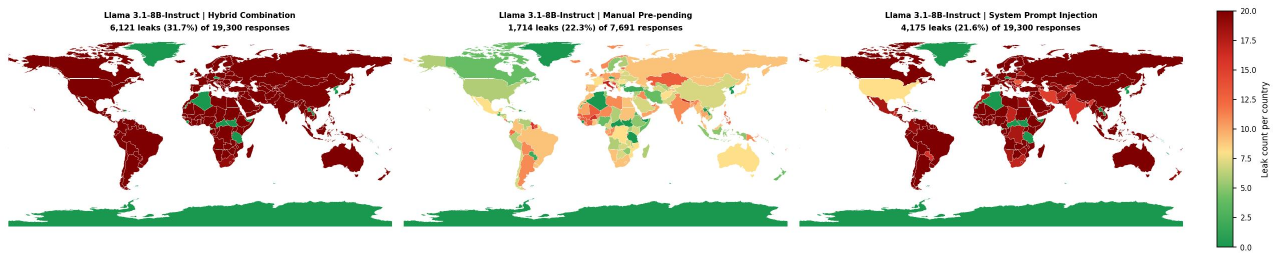}
\caption{Llama 3.1-8B-Instruct location leakage on Infinite Chats. (\textit{Left}) Hybrid: 31.7\%; (\textit{Center}) Manual Pre-pending: 22.3\%; (\textit{Right}) System Prompt Injection: 21.6\%. Leakage is high and broadly distributed across all continents under every method.}
\label{fig:llama_maps_app}
\end{figure*}

\begin{figure*}
\centering
\includegraphics[width=\textwidth]{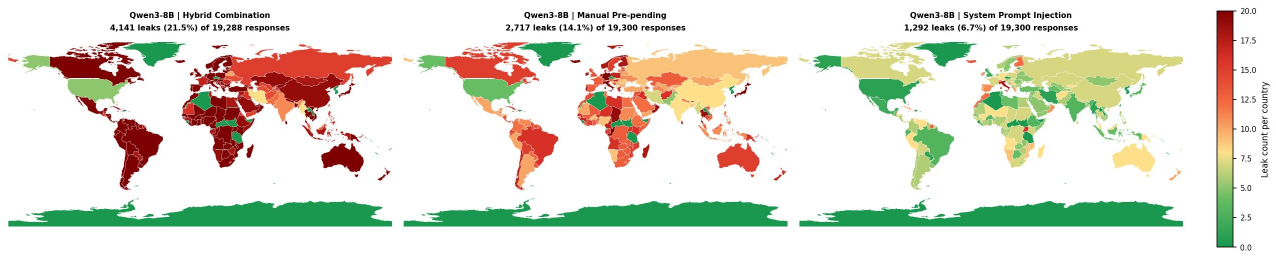}
\caption{Qwen3-8B location leakage on Infinite Chats. (\textit{Left}) Hybrid: 21.5\%; (\textit{Center}) Manual Pre-pending: 14.1\%; (\textit{Right}) System Prompt Injection: 6.7\%. Qwen shows the largest intra-model spread between methods (3.2$\times$); system-prompt injection alone produces notably sparse leakage.}
\label{fig:qwen_maps}
\end{figure*}

\begin{figure*}
\centering
\includegraphics[width=\textwidth]{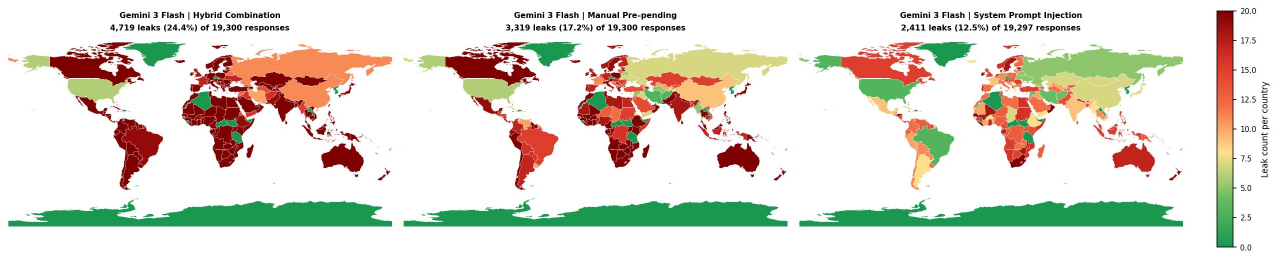}
\caption{Gemini 3 Flash location leakage on Infinite Chats. (\textit{Left}) Hybrid: 24.4\%; (\textit{Center}) Manual Pre-pending: 17.2\%; (\textit{Right}) System Prompt Injection: 12.5\%. Gemini shows broad global spread with elevated Oceania sensitivity (RSR~1.20 under Hybrid).}
\label{fig:gemini_maps}
\end{figure*}

\begin{figure*}
\centering
\includegraphics[width=\textwidth]{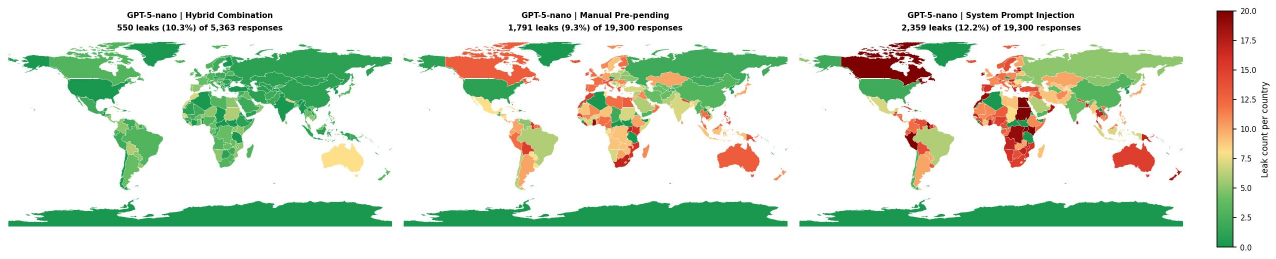}
\caption{GPT-5-nano location leakage on Infinite Chats. (\textit{Left}) Hybrid: 10.3\%$^{\dagger}$; (\textit{Center}) Manual Pre-pending: 9.3\%; (\textit{Right}) System Prompt Injection: 12.2\%. GPT-5-nano is the only model where System Prompt Injection exceeds Manual Pre-pending. $\dagger$Hybrid based on 5,363 samples.}
\label{fig:gpt_maps}
\end{figure*}

\begin{figure*}
\centering
\includegraphics[width=\textwidth]{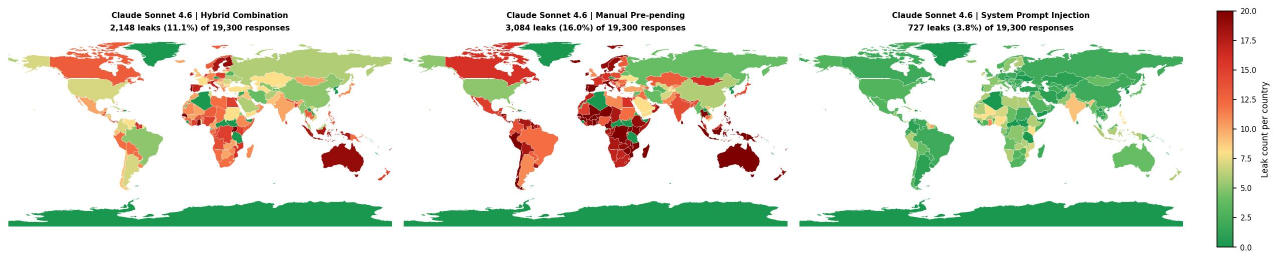}
\caption{Claude Sonnet 4.6 location leakage on Infinite Chats  (\textit{Left}) Hybrid: 11.1\%; (\textit{Center}) Manual Pre-pending: 16.0\%; (\textit{Right}) System Prompt Injection: 3.8\%. Claude exhibits the largest method asymmetry (4.2$\times$ Block vs.\ Sys) and the strongest Oceania skew of all models (RSR up to 1.62 under System Prompt Injection).}
\label{fig:claude_maps}
\end{figure*}

\autoref{fig:llama_maps}--\autoref{fig:claude_maps} show country-level leakage choropleth maps for all five models on the Infinite Chats dataset, generated from 19,300 samples ($100$ prompts $\times$ 193 countries). Each figure shows three panels: Hybrid Combination (left), Manual Pre-pending (center), and System Prompt Injection (right), on a 0--20 leak-count color scale (green = low, dark red = high).

\section{Geographic Bias Suppression via LoRA Fine-Tuning}
\label{sec:lora-finetuning}

Having established that location leakage is consistent across models and injection methods, we further ask whether it can be suppressed through fine-tuning. We fine-tuned \textit{two open-weight models Llama 3.1-8B-Instruct and Qwen3-8B using Low-Rank Adaptation (LoRA)} on a dataset where each of the 193 UN-recognized countries is paired with the same neutral, location-free target response. Therefore, if the model sees thousands of examples where different locations all map to the same output, it should learn to treat the location block as irrelevant noise. The LoRA adapter weights $\Delta\theta$ are optimized to minimize the cross-entropy loss across all geographically diverse inputs:

\begin{equation}
    \min_{\Delta\theta} \sum_{i=1}^{N} \mathcal{L}(f(x + loc_i;\, \theta + \Delta\theta_{\text{LoRA}}),\, y^*)
    \vspace{2ex}
\end{equation}

\noindent where $loc_i$ is the injected country profile for country $i$ and $y^*$ is the fixed debiased/neutral target (\autoref{sec:debiased-targets}). By holding the target constant while varying the location, this set-up explicitly penalizes the model for attending to geographic identifiers in the input, pushing it toward outputs that are consistent regardless of which country location is injected.

\subsection{Results: LoRA Fine-Tuning Pipeline}

\begin{table}[t]
\centering
\begin{tabular}{lcc}
\toprule
\textbf{Phase} & \textbf{Count} & \textbf{Rate (\%)} \\
\midrule
Pre-Fine-Tuning  & 5,303 & 13.74 $\pm$ 0.18 \\
Post-Fine-Tuning & 5,233 & 13.56 $\pm$ 0.17 \\
\midrule
$\Delta$         & $-70$ & $-0.18$ $\pm$ 0.25 \\
\bottomrule
\end{tabular}
\caption{Geographic leakage for Llama-3.1-8B-Instruct before and after fine-tuning ($N = 38{,}600$). Pre-fine-tuning: $13.74 \pm 0.18\%$; Post-fine-tuning: $13.56 \pm 0.17\%$; $\Delta = -0.18 \pm 0.25\%$ ($z = -0.73$, $p = 0.47$).}
\label{tab:llama-results}
\end{table}

As shown in \autoref{tab:llama-results}, Llama 3.1-8B exhibited a leakage rate of 13.74\% (5,303 instances) before fine-tuning. After fine-tuning, the rate dropped only marginally to 13.56\% (5,233 instances), a reduction of just 70 instances or 1.32\%. This marginal improvement shows no meaningful evidence that fine-tuning can suppress geographic conditioning.

The results were even more striking for Qwen3-8B in \autoref{tab:qwen-results}, which was fine-tuned on a larger dataset than Llama, with training samples per country extended from 200 to 500. Rather than improving, leakage actually increased from 12,350 instances pre-fine-tuning to 12,428 post-fine-tuning, a regression of $0.63\%$ ($\pm 1.18\%$, $z = +0.53$, $p = 0.60$). This suggests that Qwen3-8B's stronger pre-trained regional associations actively resisted the neutralization objective: rather than learning to ignore geographic context, the model treated the neutral canonical target as an outlier and continued to prioritize its learned regional priors. Qwen3-8B thus proved more resistant to fine-tuning than Llama3.1-8B-Instruct, amplifying leakage rather than suppressing it.

\begin{table}[t]
\centering
\begin{tabular}{lcc}
\toprule
\textbf{Phase} & \textbf{Count} & \textbf{Rate (\%)} \\
\midrule
Pre-Fine-Tuning  & 12,350 & 12.80 $\pm$ 0.11 \\
Post-Fine-Tuning & 12,428 & 12.88 $\pm$ 0.11 \\
\midrule
$\Delta$         & $+78$  & $+0.08$ $\pm$ 0.15 \\
\bottomrule
\end{tabular}
\caption{Geographic leakage for Qwen3-8B before and after LoRA fine-tuning ($N = 96{,}500$. Pre-fine-tuning: $12.80 \pm 0.11\%$; Post-fine-tuning: $12.88 \pm 0.11\%$; $\Delta = +0.08 \pm 0.15\%$ ($z = +0.53$, $p = 0.60$).}
\label{tab:qwen-results}
\end{table}

These results suggest that geographic bias is structurally ingrained in the pre-trained weights of the models and cannot be removed by lightweight post-training interventions alone. LoRA fine-tuning produced only a negligible improvement in Llama 3.1-8B-Instruct and was detrimental for Qwen3-8B, pointing to a deeper property set during pre-training rather than a surface-level behavior that fine-tuning can easily override.

\end{document}